\documentclass[letterpaper, 10 pt, conference]{ieeeconf}  

\IEEEoverridecommandlockouts                              

\usepackage{amsmath} 
\usepackage{amssymb}  
\usepackage[linesnumbered,ruled]{algorithm2e}
\usepackage{gensymb}
\usepackage{graphicx}
\usepackage{epstopdf}
\usepackage{subcaption}
\usepackage[colorlinks]{hyperref}
\usepackage{multirow}
\usepackage[T1]{fontenc}
\usepackage[super]{nth}
\usepackage{fix-cm}
\usepackage{floatrow}
\usepackage[]{caption}

\makeatletter
\let\NAT@parse\undefined
\makeatother

\usepackage[square,comma,numbers,sort&compress]{natbib} 

\title{\LARGE \bf
TACO: Trash Annotations in Context for Litter Detection
}

\author{Pedro F. Proen\c{c}a$^{\dagger}$ and Pedro Sim\~{o}es$^{\dagger}$
\thanks{$^{\dagger}$The authors worked as Independent Researchers on this project.}%
}

\begin{document}

\maketitle
\thispagestyle{empty}
\pagestyle{empty}

\begin{abstract}

TACO is an open image dataset for litter detection and segmentation, which is growing  through crowdsourcing. Firstly, this paper describes this dataset and the tools developed to support it. Secondly, we report instance segmentation performance using Mask R-CNN on the current version of TACO.
Despite its small size (1500 images and 4784 annotations), our results are promising on this challenging problem. However, 
to achieve satisfactory trash detection in the wild for deployment, TACO still needs much more manual annotations. These can be contributed using: {\small \url{http://tacodataset.org/}}

\end{abstract}


\section{Introduction}

Litter\footnote{Any manufactured solid material that is incorrectly disposed either intentionally or accidentally.} has been accumulating around us as most local governments and international organizations fail to tackle this crisis, which is having a catastrophic impact on biodiversity and marine animals \cite{derraik2002pollution,gall2015impact}. While littering is widely considered illegal, effective means to control it are currently lacking, both technological and regulatory. \par 
Fortunately, there has been an increasing number of initiatives \cite{maximenko2019towards} to implement litter monitoring systems from remote sensing to \textit{in situ} observations. We believe these systems need a high degree of autonomy enabled by deep learning. For that, we need annotated photos of litter in context, as in \cite{chiba2018human}, to train an evaluate litter detectors -- rather than clean images of trash with white background \cite{awe2017smart}. Despite the current availability of large general image datasets, trash is poorly represented. \par
Detecting trash in the wild can be a very challenging problem -- more than trash in recycling facilities, e.g., conveyor belt, bins. Not only do we have to take into account that trash can be deformable, transparent, aged, fragmented, occluded and camouflaged, we also need models to be aware of the vast diverse features that make our natural world.
With this in mind, this work introduces TACO, an effort to build a comprehensive dataset of photos taken from diverse environments around the word (e.g. beaches, cities) with litter segmented and annotated using a hierarchical taxonomy. The next section describes its main features, current stats and supplementary tools. Then, Section \ref{cap:experiments} presents our litter detection experiments and discusses results on two different tasks with this new dataset.

\begin{figure}[t]
\centering
	\begin{tabular}{@{}c@{ }c@{ }c@{}}
		\includegraphics[width=.32\linewidth]{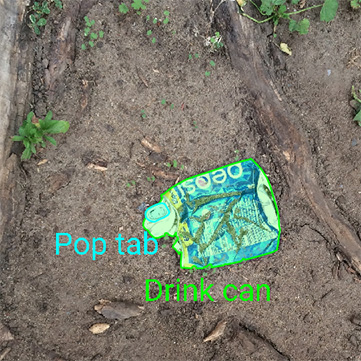} &
		\includegraphics[width=.32\linewidth]{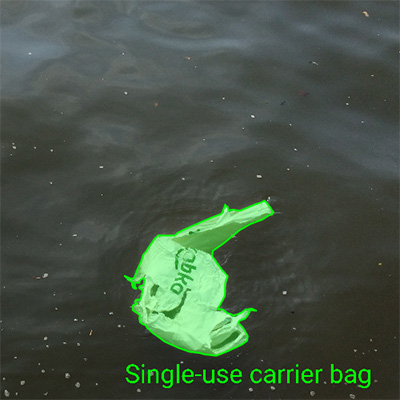}&
		\includegraphics[width=.32\linewidth]{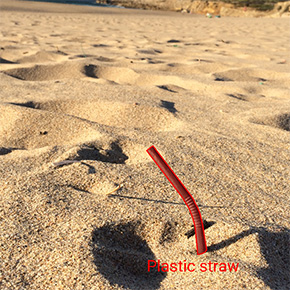} \\
		\includegraphics[width=.32\linewidth]{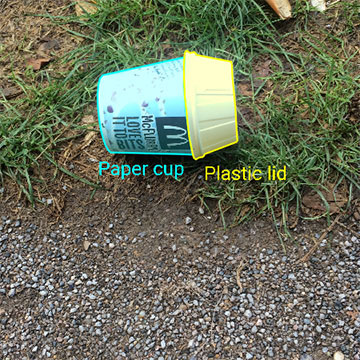} &
		\includegraphics[width=.32\linewidth]{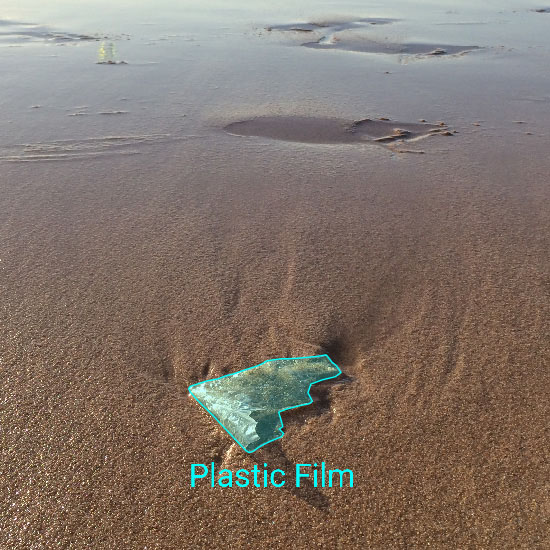}&
		\includegraphics[width=.32\linewidth]{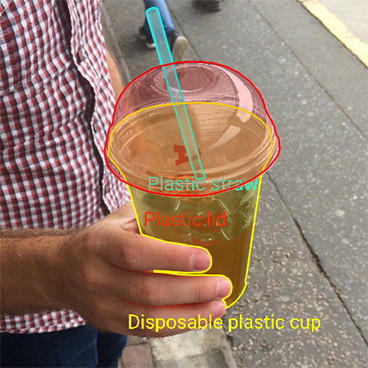}\\
		\includegraphics[width=.32\linewidth]{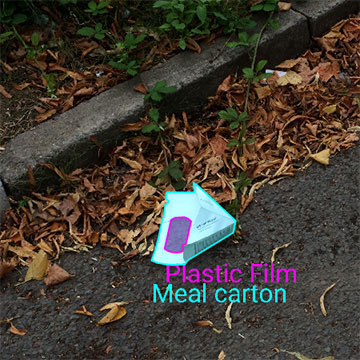} &
		\includegraphics[width=.32\linewidth]{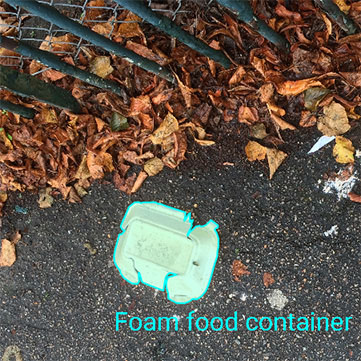}&
		\includegraphics[width=.32\linewidth]{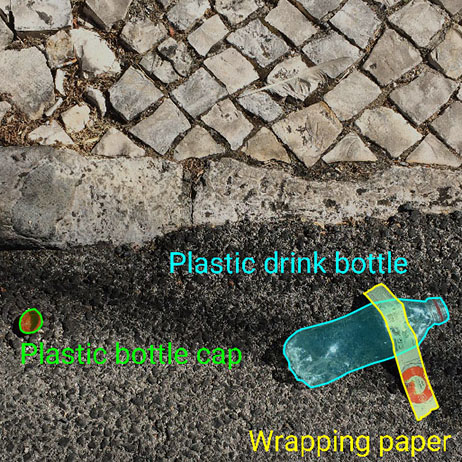}
	\end{tabular} 
	\caption{Cropped annotated images from TACO dataset}
	\label{fig:teaser}
\end{figure}

\section{TACO dataset}

\begin{figure}[t]
	\centering
	\includegraphics[scale=0.5]{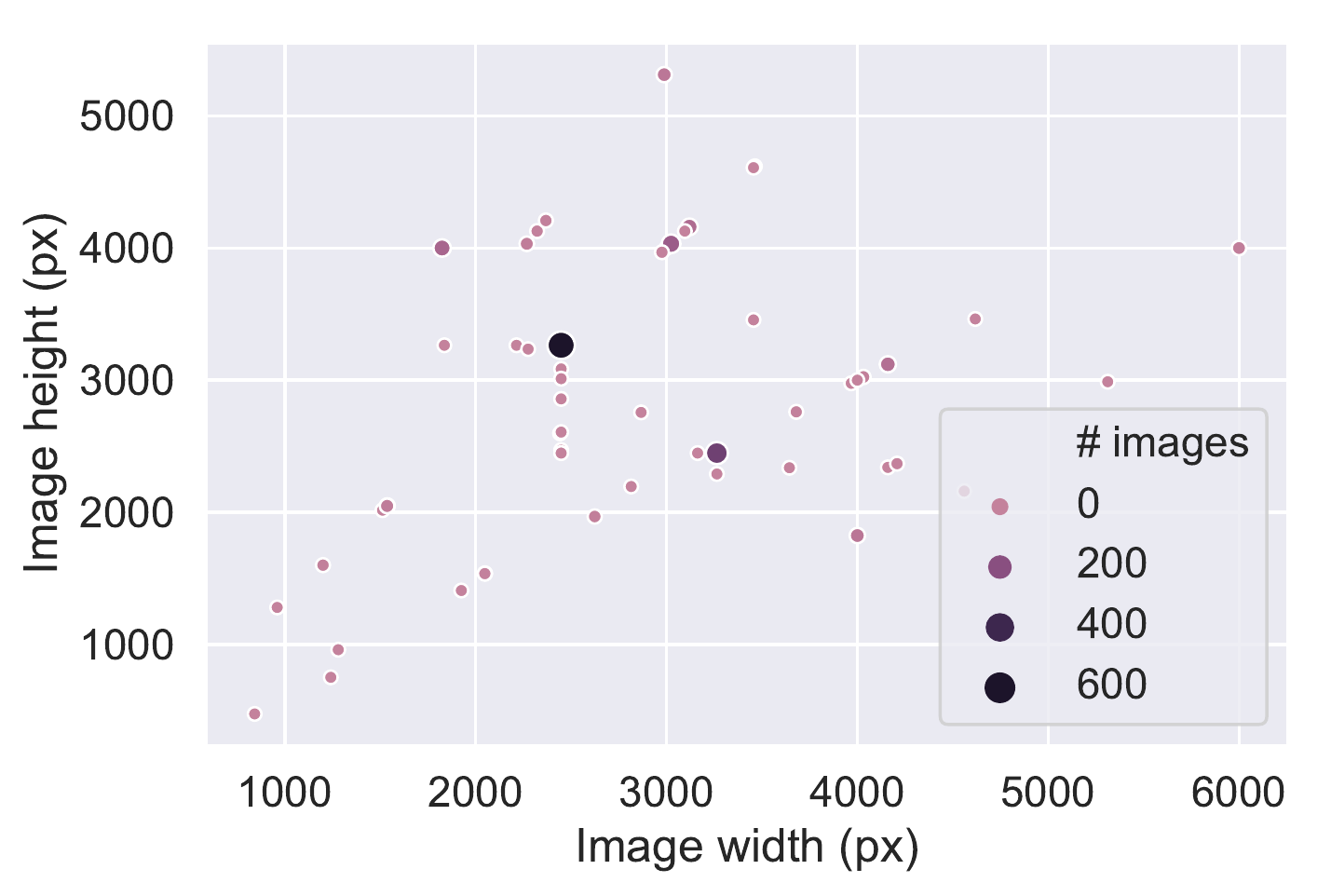}
	\caption{Distribution of image resolutions.}
	\label{fig:img_resolutions}
\end{figure}

TACO contains high resolution images, as shown in Fig. \ref{fig:img_resolutions}, taken mostly by mobile phones. These are managed and stored by Flickr, whereas our server manages the annotations and runs periodically a crawler to collect more potential images of litter. Additionally, we also selected some images from \cite{Openlittermap}. All images are under free copyright licences and are annotated and segmented by users using our online tool: {\small \url{http://tacodataset.org/annotate}}.\par

\begin{figure}
	\centering
	\includegraphics[scale=0.48]{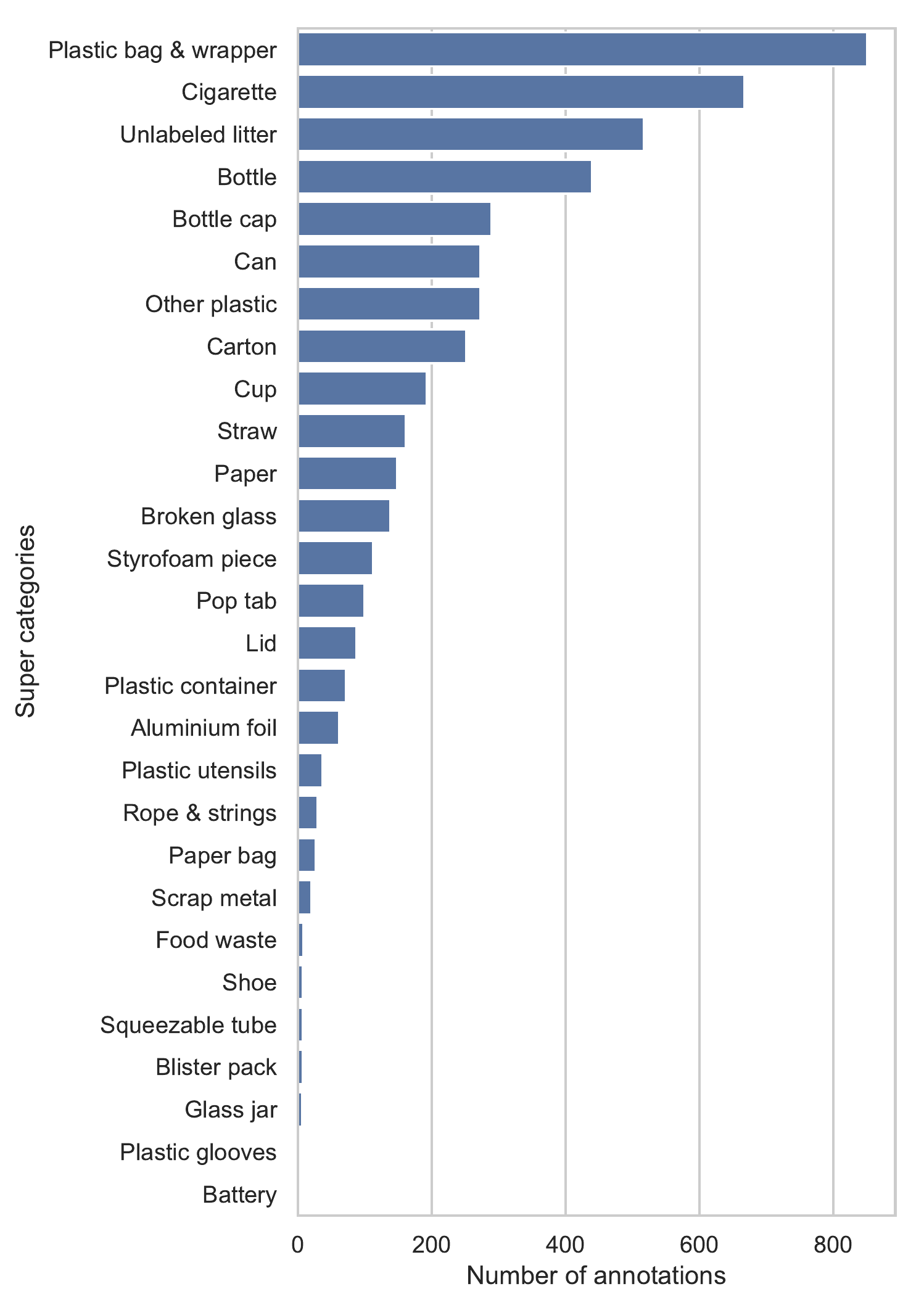}
	\caption{Number of annotations per super category provided by TACO's current version.}
	\label{fig:cat_hist}
\end{figure}

\begin{figure}
	\centering
	\includegraphics[scale=0.4]{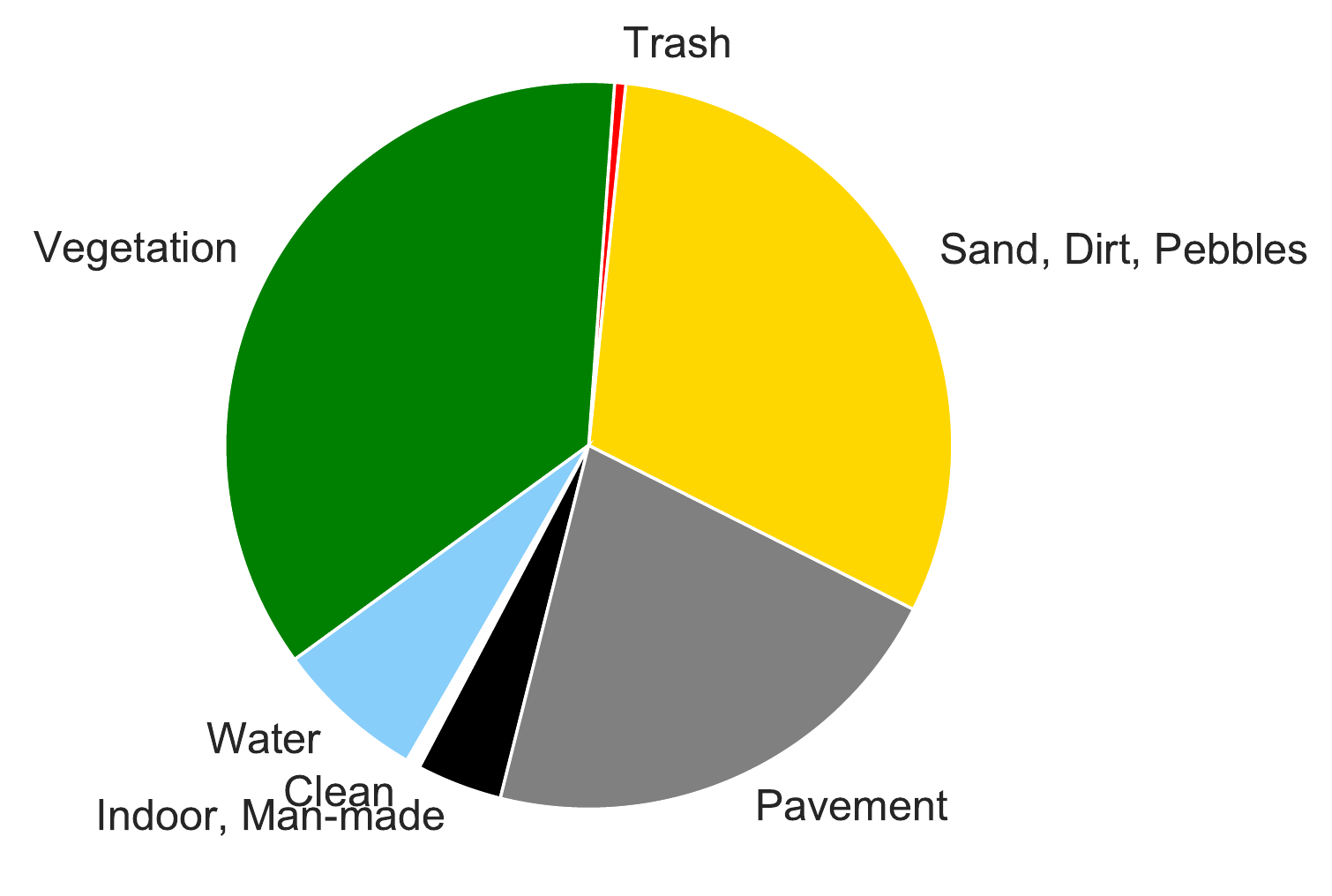}
	\caption{Proportion of images by background tag.}
	\label{fig:background}
\end{figure}

\begin{figure}[t]
	\centering
	\includegraphics[scale=0.5]{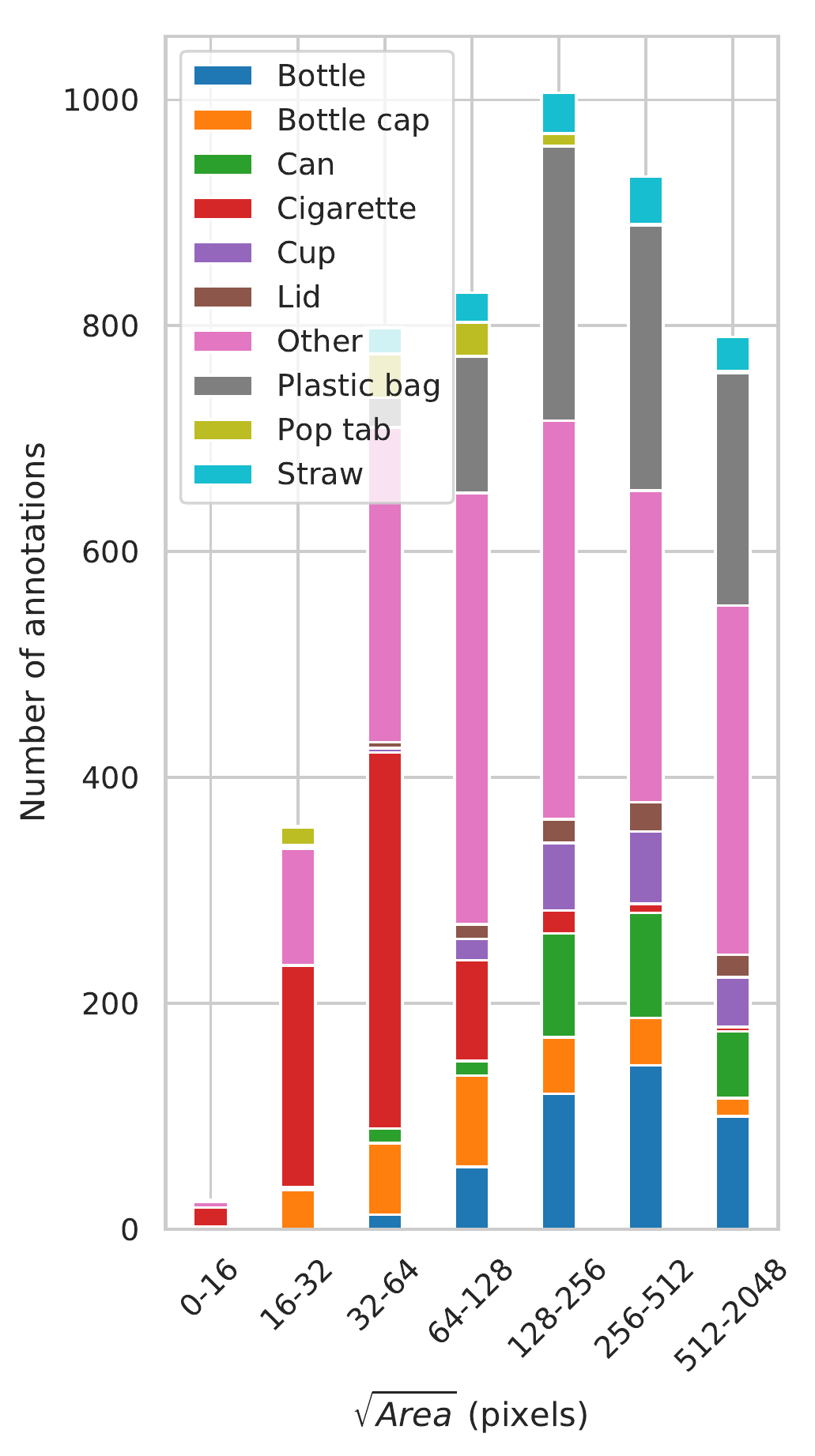}
	\caption{Histogram of bounding box sizes per category for TACO-10.}
	\label{fig:area_per_category}
\end{figure}

 Specifically, images are labeled with the scene tags, shown in Fig. \ref{fig:background}, to describe their background -- these are not mutually exclusive -- and litter instances are segmented and labeled using a hierarchical taxonomy with 60 categories of litter which belong to 28 super (top) categories, shown in Fig. \ref{fig:cat_hist}, including a special category: \textit{Unlabeled litter} for objects that are either ambiguous or not covered by the other categories. This is fundamentally different from other datasets (e.g. COCO) where distinction between classes is key. Here, all objects can be in fact classified as one class: \textit{litter}\footnote{In this work we assume all annotated objects to be litter, even though there are a small number of objects that are handheld, not disposed yet or their location context is ambiguous.} . Furthermore, it may be impossible to distinguish visually between two classes, e.g., plastic bottle and glass bottle. Given this ambiguity and the class imbalance shown in Fig. \ref{fig:cat_hist}, classes can be rearranged to suit a particular task. For our experiments, in Section \ref{cap:experiments}, we targeted 9 super categories based on the number of instances and merged the rest under the class name \textit{Other Litter}. We call this TACO-10 and Fig. \ref{fig:area_per_category} shows the size variability of annotations per category for this new taxonomy. We can see that most of the cigarettes, the third largest class, have an area less than $64\times64$ pixels. We will see in our results how this can be problematic.

\subsection{Litter transplants}

\begin{figure}
	\centering
	\begin{tabular}{@{}c@{ }c@{}}
		\includegraphics[width=.45\linewidth]{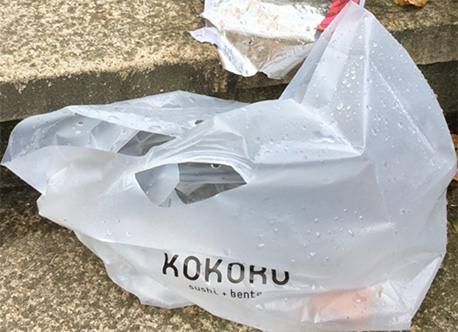} &
		\includegraphics[width=.465\linewidth]{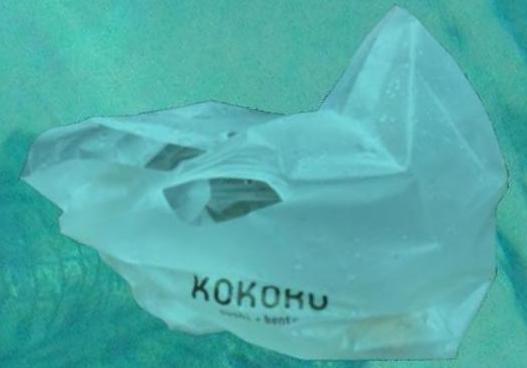}\\
		\includegraphics[width=.45\linewidth, height=82pt]{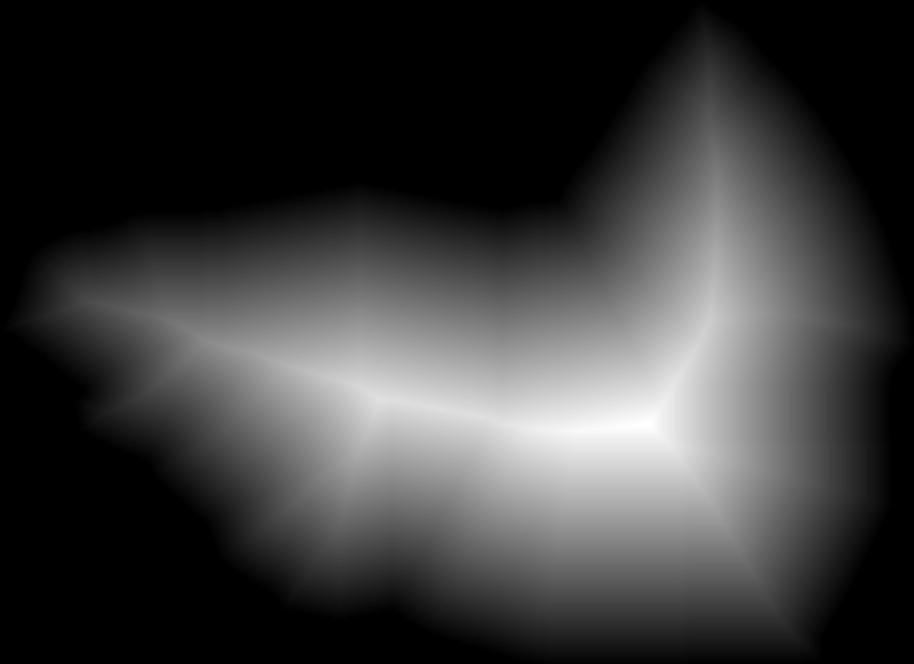}&
		\includegraphics[width=.465\linewidth]{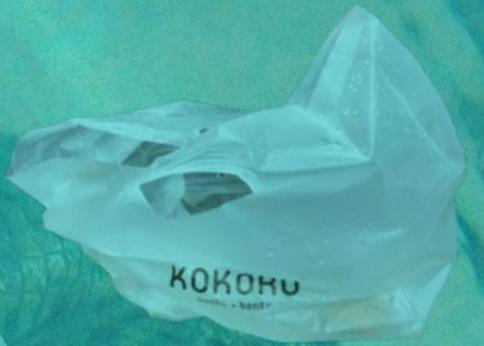}  \\
	\end{tabular}
	\caption{Soft vs hard transplant. \textit{Top-left}: Original object. \textit{Top-right}:  Object transplanted to new image using original segmentation mask. \textit{Bottom-right}: Object transplanted to new image using instead a soft mask obtained via distance transform, shown on \textit{Bottom-left}. Notice the silhouette is more even.}
	\label{fig:transplants}
\end{figure}

Transplantation can be an effective way of augmenting the dataset by adding more under-represented backgrounds if the mask segmentations are precise enough. This is especially relevant if we are going to operate on closed-set conditions (i.e. specific environment). Suppose we want to run litter detection on a river with crocodiles and make sure those crocodiles do not give rise to false positives, but this river does not have a lot of litter to take pictures add to our dataset, and we sure do not want to interfere with their habitat! What one could do is simply copy paste TACO segmentations onto images with crocodiles as shown in image \ref{fig:transplants}, but this will create artifacts around the edges because the masks are not perfect. Thus, as illustrated in Fig. \ref{fig:transplants}, we propose to pixelwise mix the transplanted object and the target image by using a truncated distance transform of the binary mask, that is a smooth mask that effectively smooths the silhouette.
 
\begin{figure}[h]
	\centering
	\includegraphics[scale=0.51]{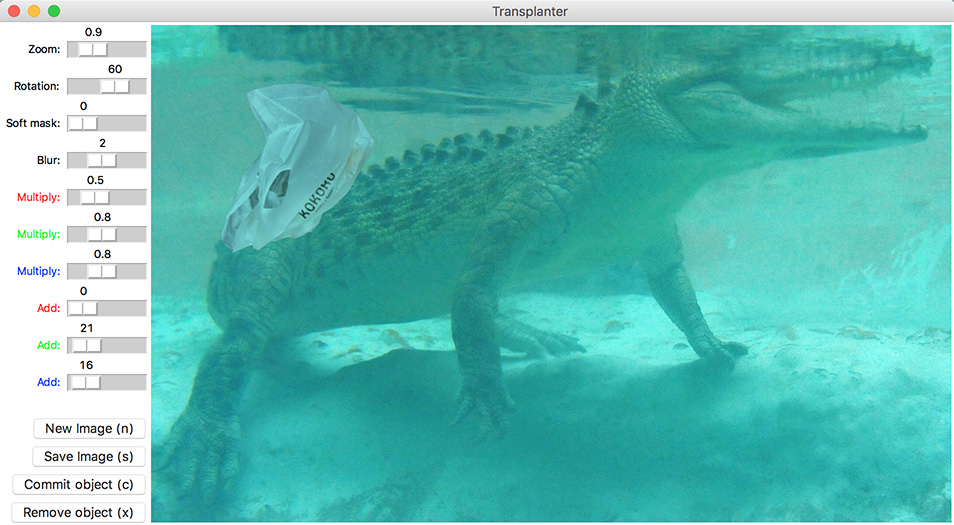}
	\caption{Transplanter. Our proposed GUI to transplant dataset segmentations.}
	\label{fig:transplanter}
\end{figure}

This function is embedded in our proposed GUI, shown in Fig. \ref{fig:transplanter}, along with other features to make transplantation easy and seamless. It's worth noting this does not work well with transparent objects (e.g. plastic films) and although this can make transplantation look natural, it neglects many lighting and camera aspects to make it realistic.

\section{Experiments}
\label{cap:experiments}

To assess the performance of litter detection and segmentation using TACO, we evaluated Mask R-CNN on two separate tasks (i.e. taxonomies): (i) TACO-1, classless litter detection where the goal is to only detect and segment litter items; and (ii) TACO-10, detection and classification to distinguish these 10 classes of litter.

Due to the small size of this dataset. All quantitative results are reported for a 4-fold cross validation. For each fold, the dataset is split randomly into 80\% for training set, 10 \% for validation set and 10 \% for test set. As an evaluation metric, we rely on the established Average Precision (AP) \cite{everingham2010pascal}  averaged over Intersection-over-Union (IoU) thresholds using the instance masks, and tested different scores to rank predictions as described in Section \ref{sec:scoring}.  The next subsection describes our Mask R-CNN implementation and training details. 

\subsection{Implementation}
We adopted the Mask R-CNN implementation by \cite{matterport_maskrcnn_2017}. Our adapted implementation and trained model weights used in this work are publicly available at: {\small \url{https://github.com/pedropro/TACO}}. We used simply the default Resnet-50 in a Feature Pyramid Network as a backbone with an input layer size of 1024$\times$1024 px by resizing and padding images accordingly.  Models were trained on TACO-10 with SGD for 100 Epochs with a batch size of 2 images and a learning rate of 0.001. Weights were started using Mask R-CNN weights trained on COCO dataset \cite{lin2014microsoft}. Longer training continues to reduce the training error but does not seem to reduce the validation error. For data augmentation, during training, we added Gaussian blur and AWG noise, changed image exposure and contrast, rotated images between $[-45^\circ, 45^\circ]$ and cropped images around the annotated bounding boxes, such that there is always a visible litter object.
To further augment the training set, we also transplanted 320 instances from the training set to images crawled from Flickr using tags based on the the scene tags shown in Fig. \ref{fig:background}.

\subsection{Prediction Scoring}
\label{sec:scoring}
As demonstrated in \cite{huang2019mask}, AP depends significantly on the score used to rank the predictions, and the established maximum class confidence may not the best choice. Therefore, to suit our two tasks, we tested using 3 different scores from the output of the Mask R-CNN classification \textit{head}. Let the class probabilities, given by this \textit{head}, be $P=\{p_1, p_2, ..., p_{N+1}\}$ where $N$ is the number of classes and $p_{N+1}$ is the probability of being background, then we compared using the following scores:

\begin{equation}
\label{eq:loc_total}
Score =\left\{
\begin{array}{rr}
\max_{\,i} p_i, \quad class\_score\\ \\
1- p_{N+1}, \quad litter\_score\\ \\ 
\frac{\max_{\,i}  p_i}{p_{N+1}+\epsilon}, \quad ratio\_score\\
\end{array}
\right.
\end{equation}
\vspace{10pt}

While \textit{class\_score} is the established score, \textit{ratio\_score} expresses both the confidence on a class and the confidence on being litter, where $\epsilon$ is just a small scalar to avoid NaN. That is, \textit{ratio\_score} allows us to say "This model is \textit{ratio\_score} times more confident that given object is class X than not being litter."

\begin{table}[b]
	\centering
	\footnotesize{
		\begin{tabular}{llll}
			\hline
			Dataset & Class score & Litter score & Ratio score \\
			\hline
			TACO\_1 & 15.9 $\pm$ 1.0 & 26.2  $\pm$ 1.0 & 26.1 $\pm$ 1.0 \\
			TACO\_10 & 17.6 $\pm$ 1.6 & 18.4  $\pm$ 1.5 & 19.4  $\pm$ 1.5 \\
			\hline
		\end{tabular}
	}
	\caption{AP for 4-fold cross validation using different scores to rank predictions.}
	\label{tab:final_AP}
\end{table}

\begin{figure}[t]
	\centering
	\begin{tabular}{@{}c@{}}
		\includegraphics[width=0.8\linewidth]{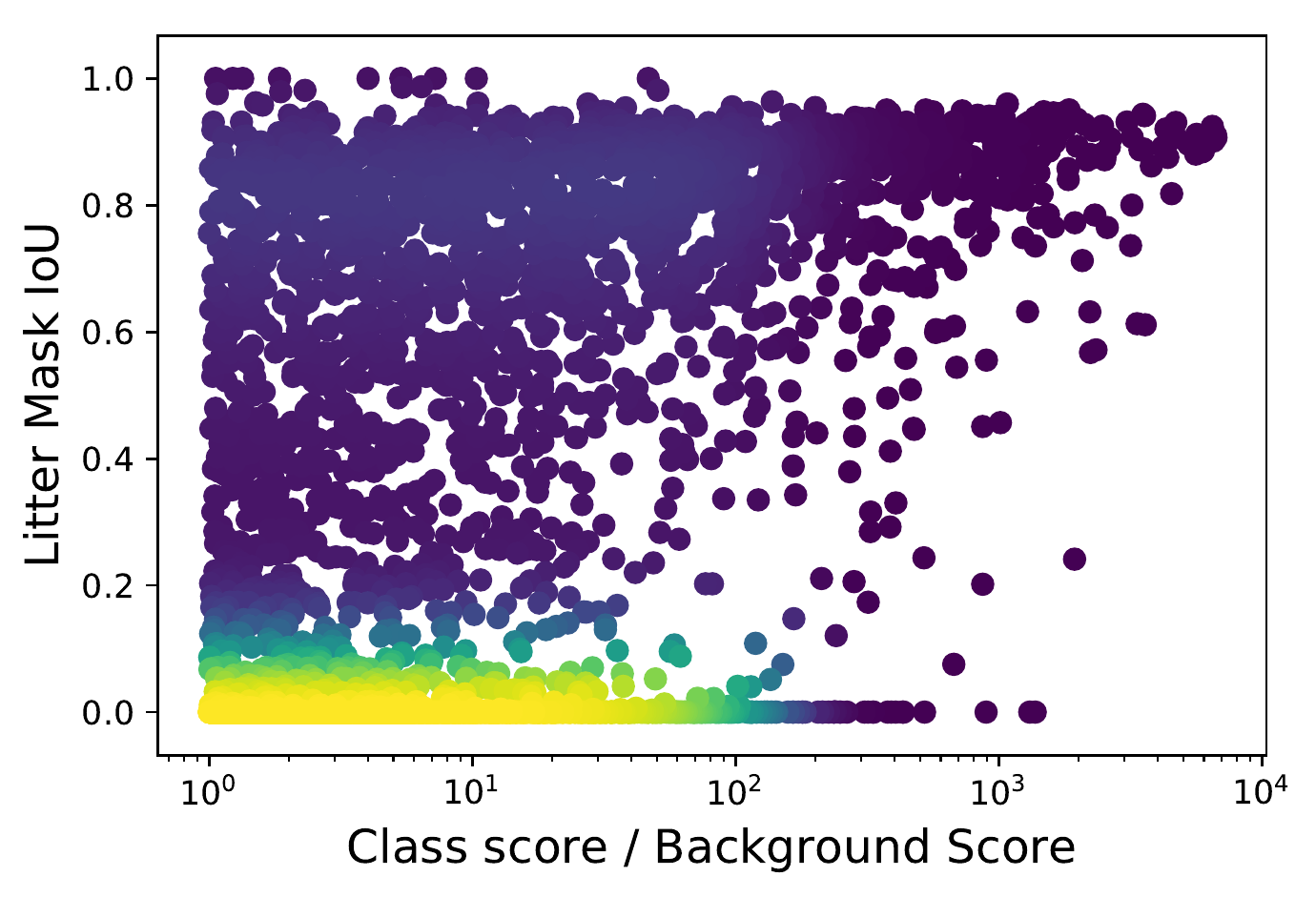}\\
		\includegraphics[width=0.8\linewidth]{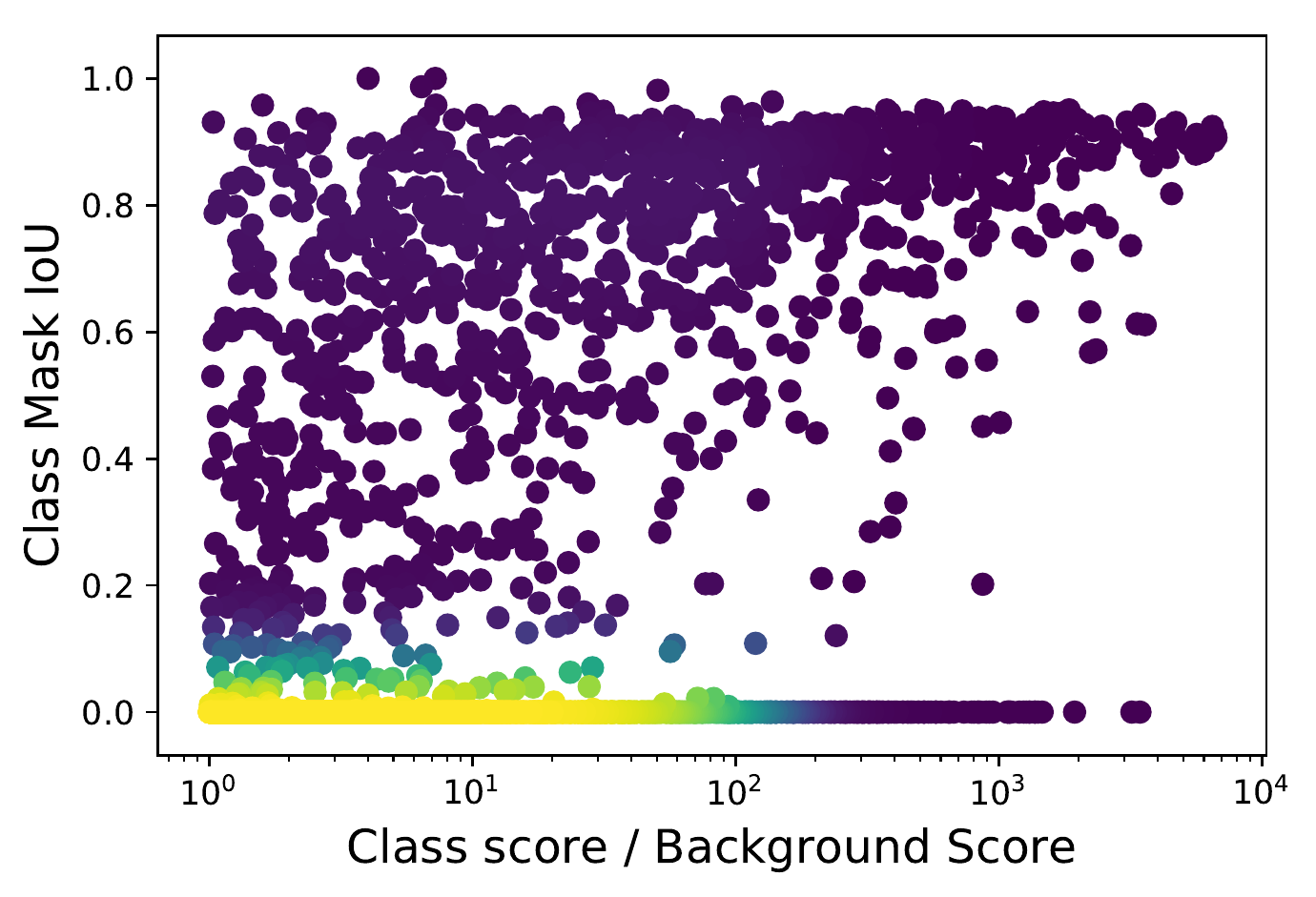}\\
	\end{tabular} 
	\caption{Segmentation IoU vs ratio score for all detection instances obtained from 4 testing sets. \textit{Top}: IoU is obtained for best classless matches (TACO-1). \textit{Bottom}: IoU is obtained  for best TACO-10 matches. Color represents density of scattered points.}
	\label{fig:IOUs}
\end{figure}

\begin{figure*}[h]
	\centering
	\begin{tabular}{@{}c@{}c@{}}
		\includegraphics[width=0.5\linewidth]{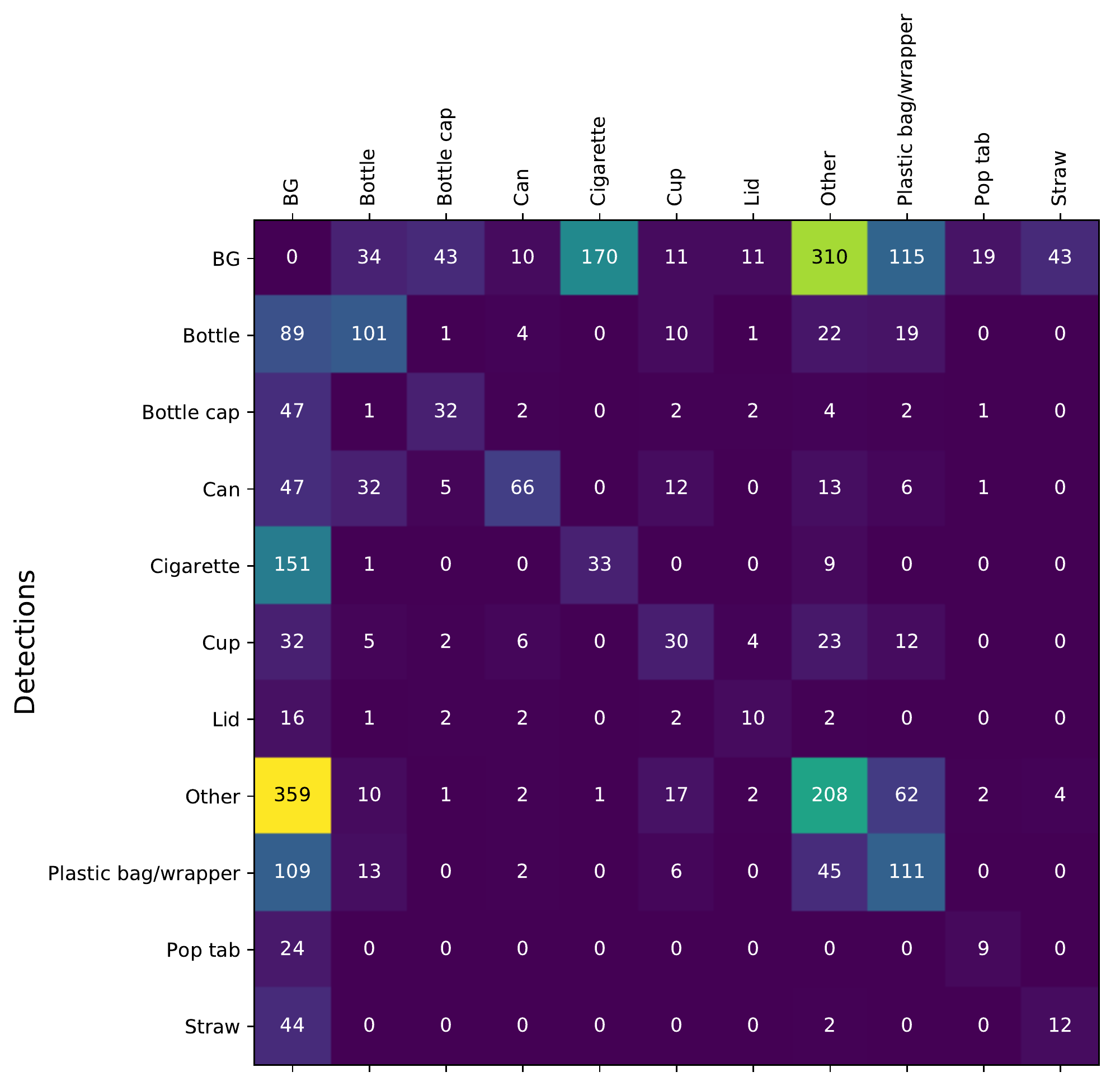}&
		\includegraphics[width=0.5\linewidth]{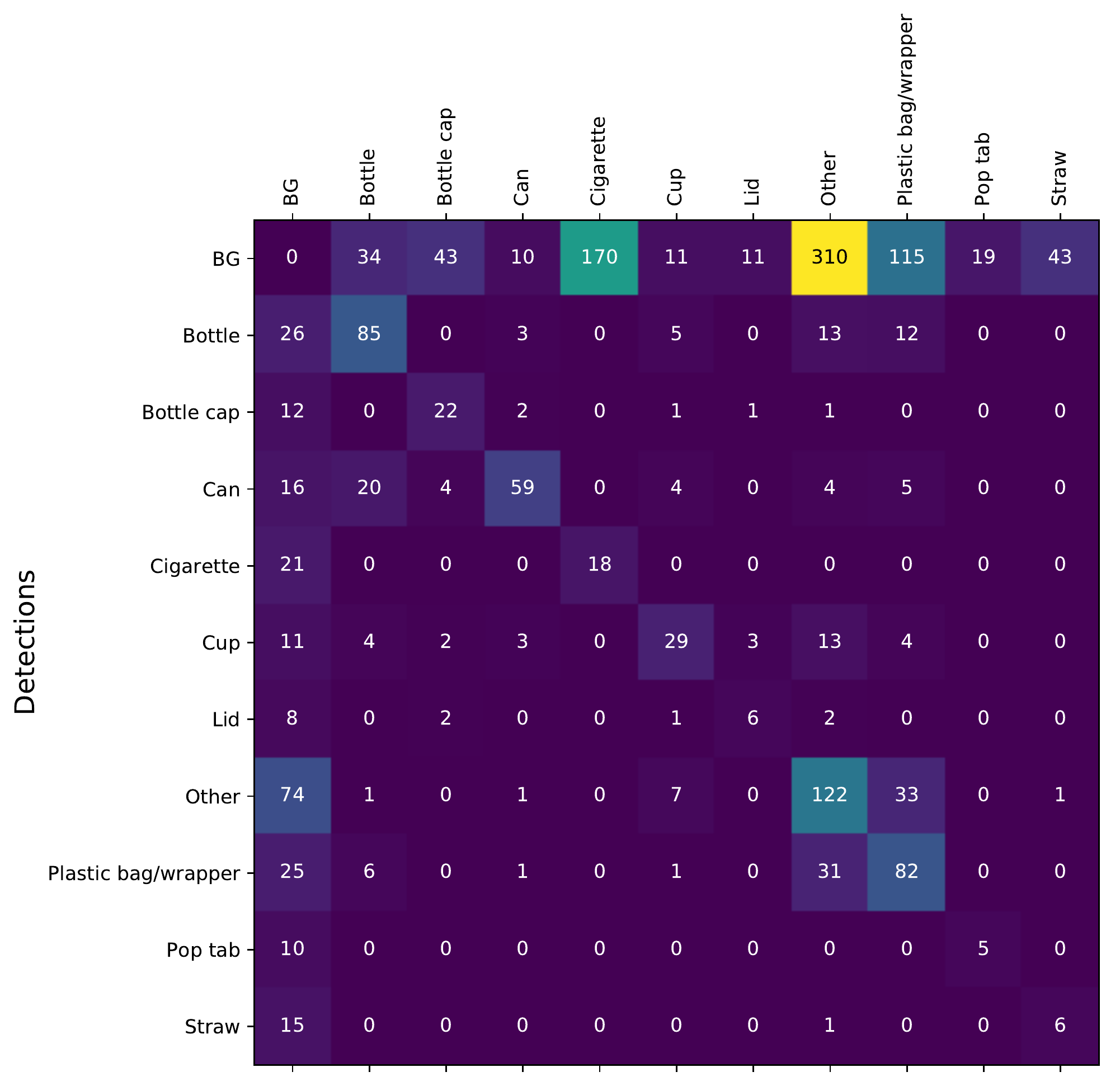}\\
	\end{tabular} 
	\caption{Confusion matrices for matches with IoU>0.5. (Left) detections with \textit{ratio score}>10, (Right) detections with \textit{ratio score}>50. \textit{BG} represents background, thus false positives are counted on the first column and false negatives on the first row.}
	\label{fig:confusion_matrices}
\end{figure*}

\begin{figure*}[h]
	\centering
	\includegraphics[scale=0.5]{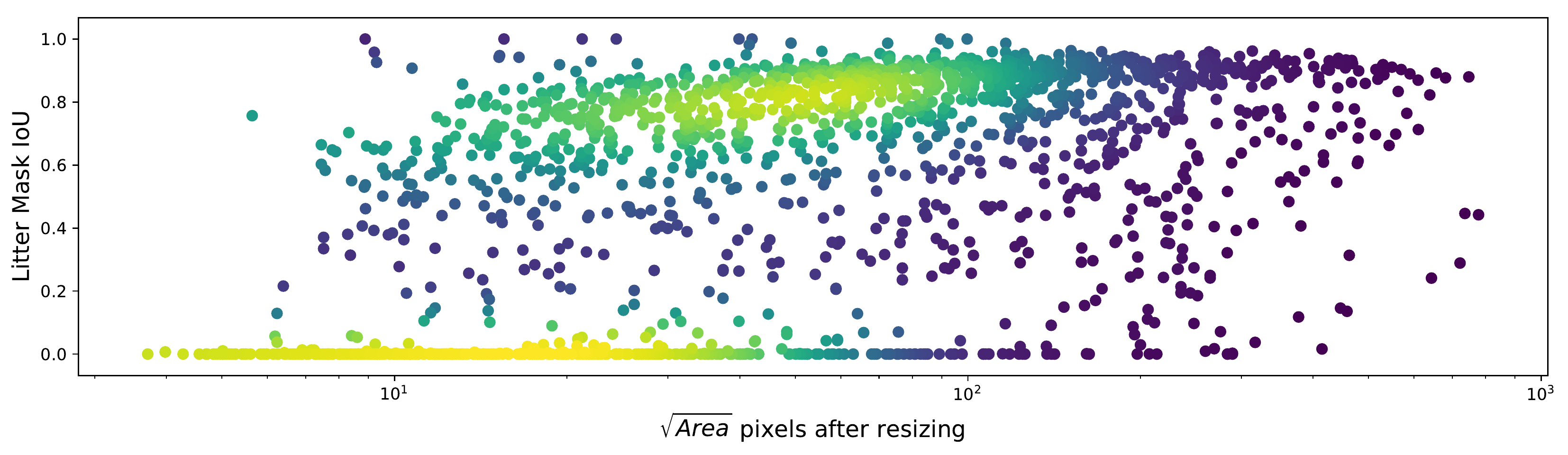}
	\caption{Annotation bounding box area vs best matched IoU. Color represents density of scattered points.}
	\label{fig:area_vs_iou}
\end{figure*}

\subsection{Results}

Table \ref{tab:final_AP} reports the AP for our proposed two tasks using different scores to rank the predictions.
As expected, the typically used \textit{class\_score} is not suitable to rank classless litter detections on TACO-1 compared to using the $\textit{litter\_score}= 1-\textit{background\_score}$. More interestingly,  \textit{class\_score} is not more reliable than \textit{litter\_score} on the TACO-10 problem, whereas 
using \textit{ratio\_score} actually improves the AP on TACO-10 and does not decrease the AP on TACO-1. By expressing this relation; How confident I am that this object is this class of litter and not background? we suit simultaneously both tasks. As we can see in Fig \ref{fig:IOUs}. this score correlates well with the Intersection-over-Union (IoU) of the predictions. 
\par
Fig. \ref{fig:confusion_matrices} shows the respective confusion matrices for TACO-10. We can conclude that our low performance in Table \ref{tab:final_AP} is mostly due to poor cigarette detection which exhibits high false positives and negatives. We believe this is due to their small size, as shown in Fig. \ref{fig:area_per_category}, since most images had to be resized to less than one third (1024$\times$1024 px). In fact, we can clearly see, in Fig. \ref{fig:area_vs_iou}, that a large number of ground truth objects with less than 20$\times$20 px that are missed.  \par

On the other hand, we also see in Fig.  \ref{fig:confusion_matrices} that the detection performance is better with Cans and Bottles although a significant number of bottles are detected as cans. It is also worth noting that there is some confusion between \textit{Plastic bag} and \textit{Other}, which is not surprising if we consider \textit{Other} includes objects with similar materials. \par

\begin{figure*}
	\centering
	\begin{tabular}{@{}c@{ }c@{ }c@{ }c@{}}
		Predicted & Groundtruth & Predicted & Groundtruth \\
		\includegraphics[width=0.245\linewidth]{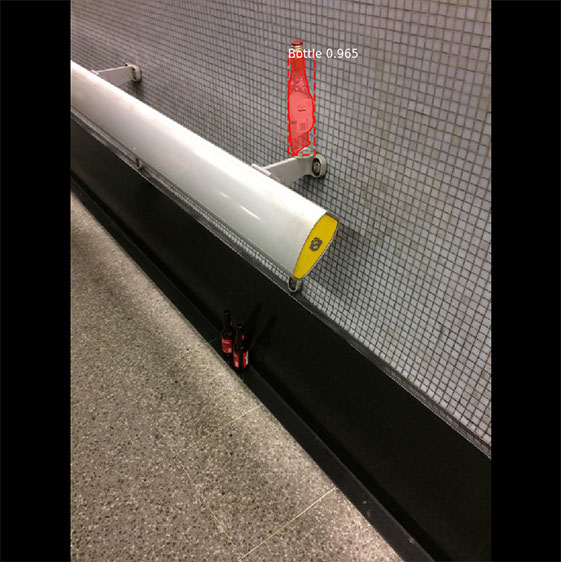}&
		\includegraphics[width=0.245\linewidth]{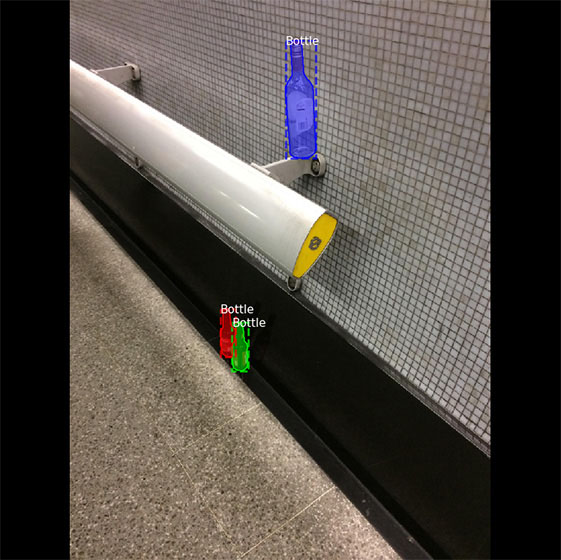}&
		\includegraphics[width=0.245\linewidth]{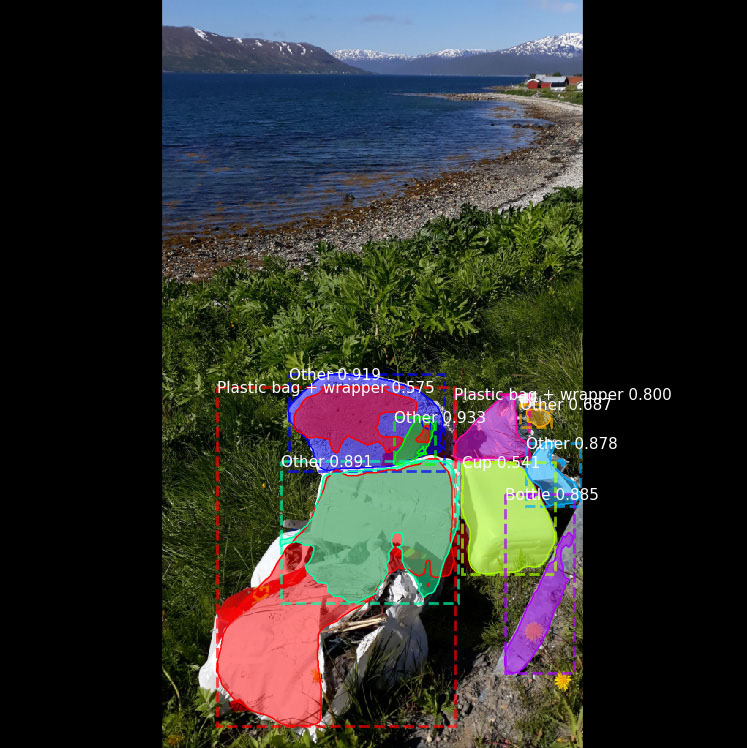}&
		\includegraphics[width=0.245\linewidth]{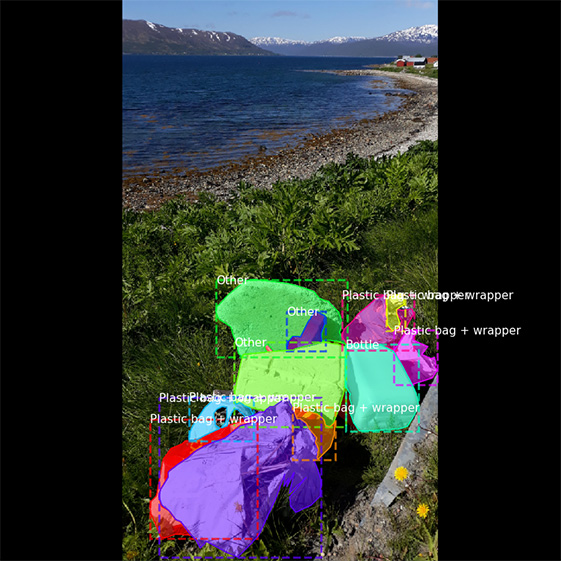}\\
	\end{tabular} 
	\caption{Examples of Mask R-CNN predictions on TACO-10 test-sets.}
	\label{fig:more_ex}
\end{figure*}

\begin{figure*}
	\centering
	\begin{tabular}{@{}c@{ }c@{ }c@{ }c@{}}
		Predicted & Groundtruth & Predicted & Groundtruth \\
		\includegraphics[width=0.245\linewidth]{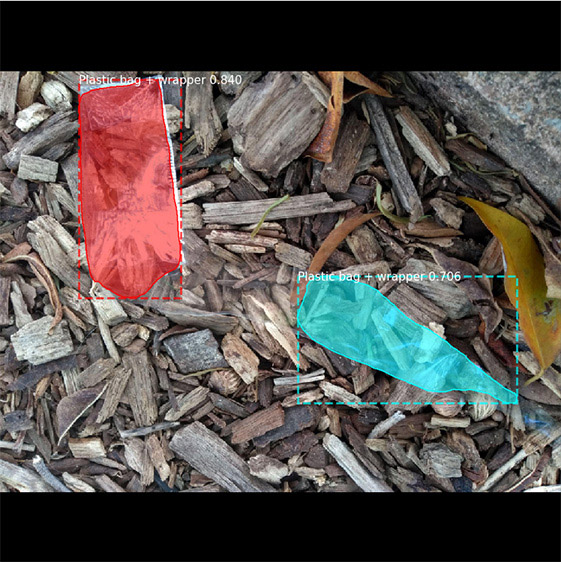}&
		\includegraphics[width=0.245\linewidth]{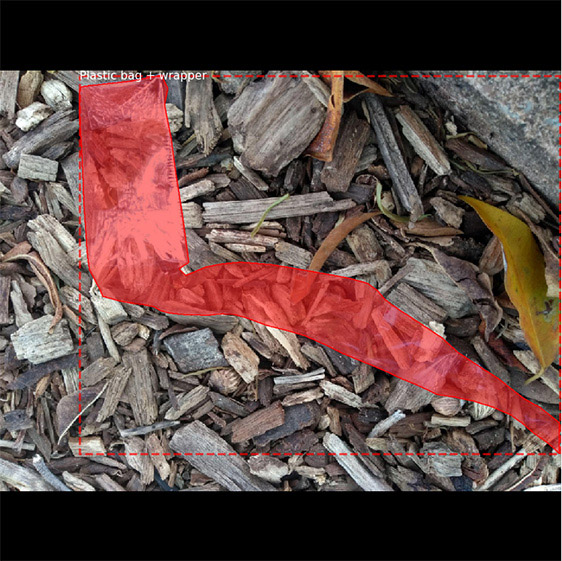}&
		\includegraphics[width=0.245\linewidth]{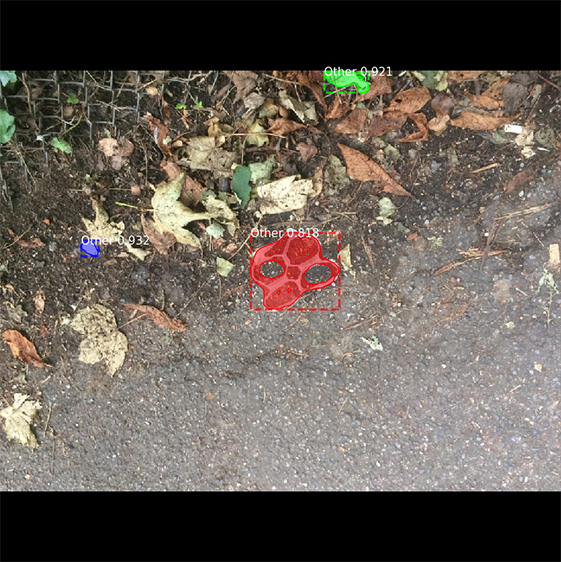}&
		\includegraphics[width=0.245\linewidth]{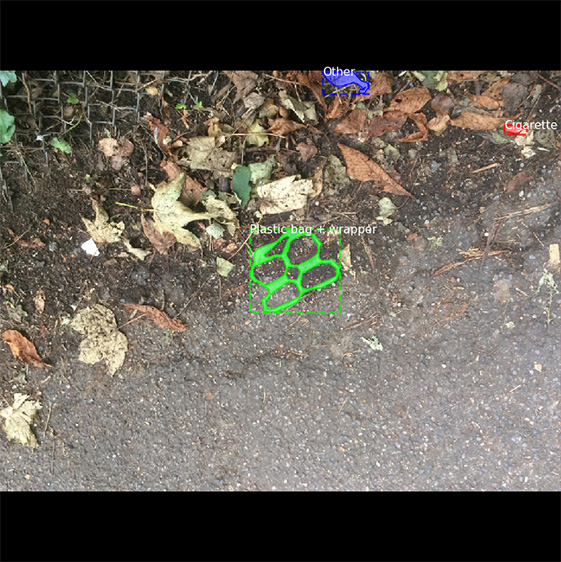}\\
		\includegraphics[width=0.245\linewidth]{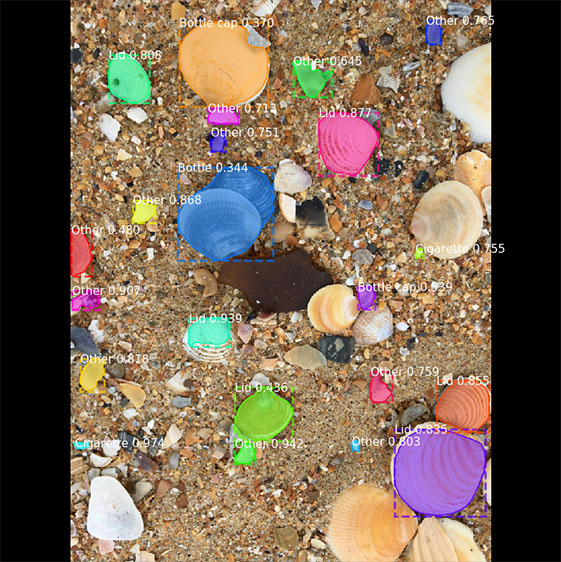}&
		\includegraphics[width=0.245\linewidth]{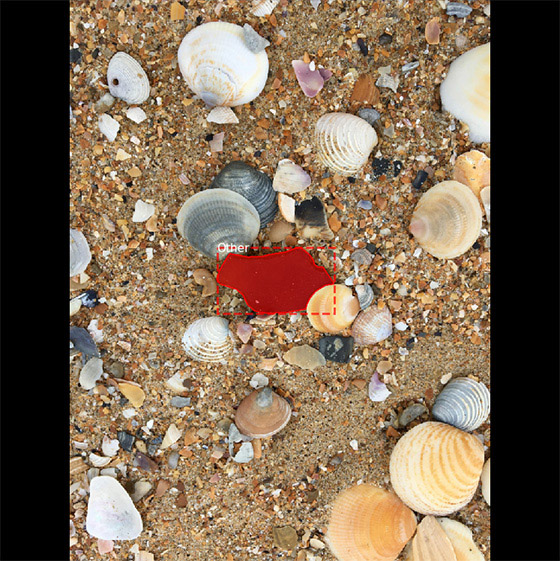}&
		\includegraphics[width=0.245\linewidth]{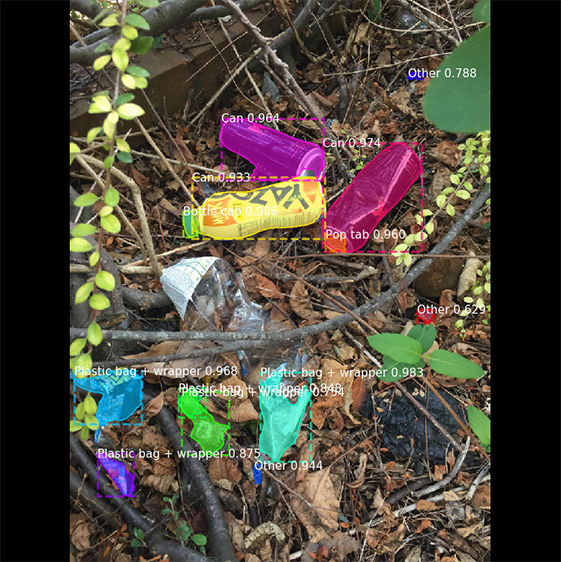}&
		\includegraphics[width=0.245\linewidth]{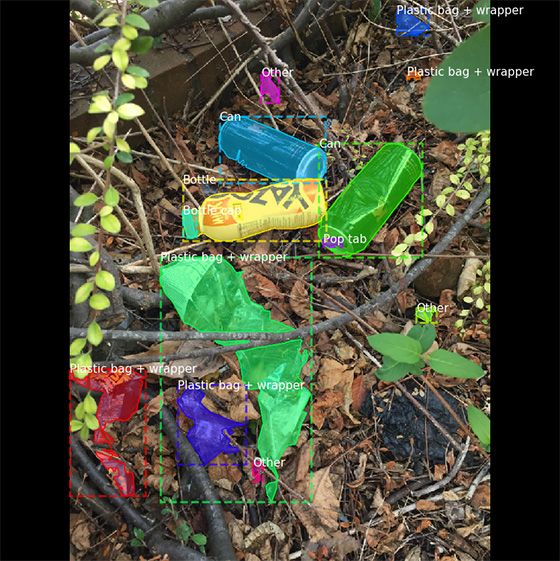}\\
		\includegraphics[width=0.245\linewidth]{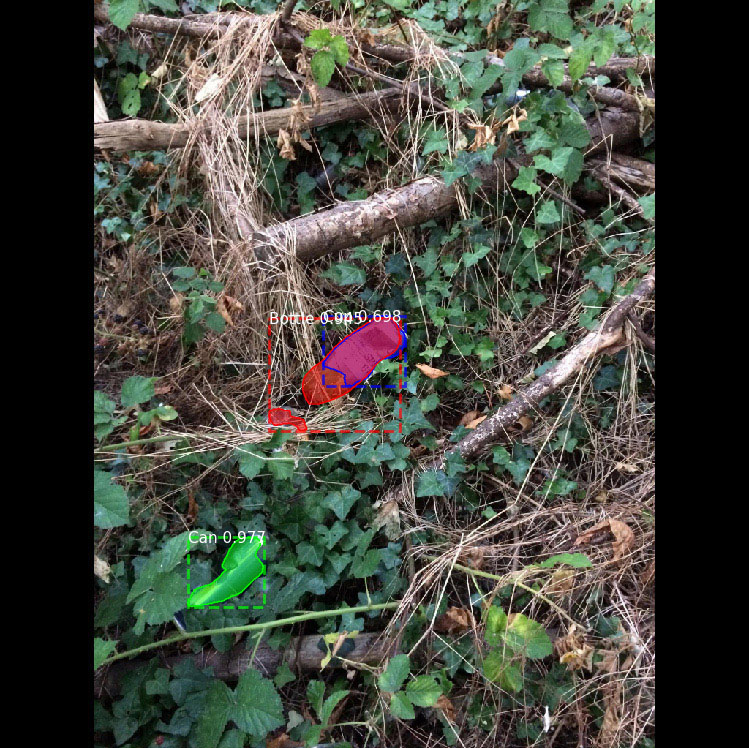}&
		\includegraphics[width=0.245\linewidth]{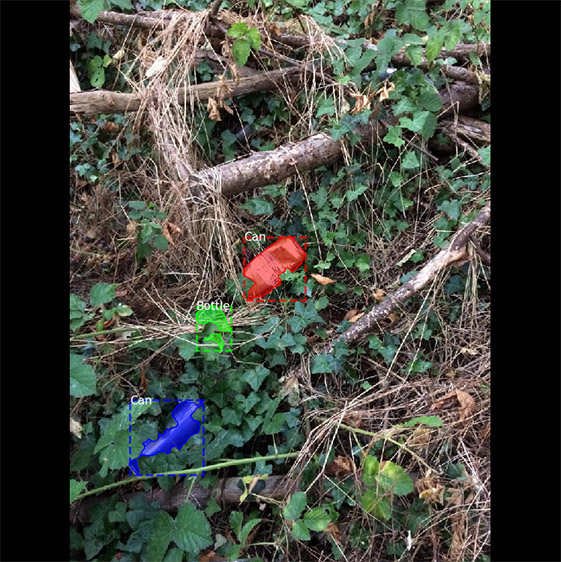}&
		\includegraphics[width=0.245\linewidth]{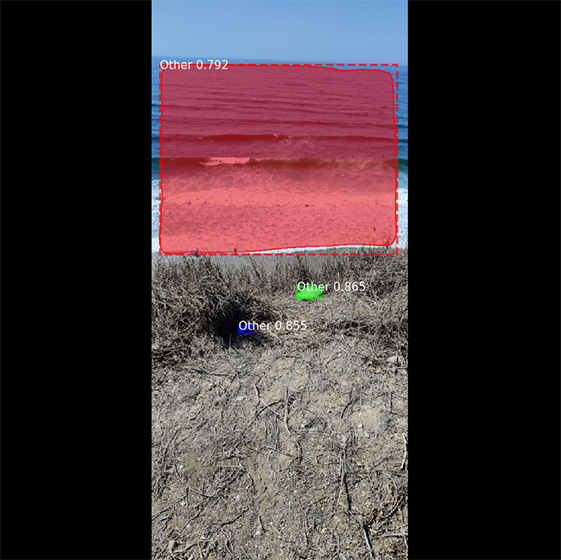}&
		\includegraphics[width=0.245\linewidth]{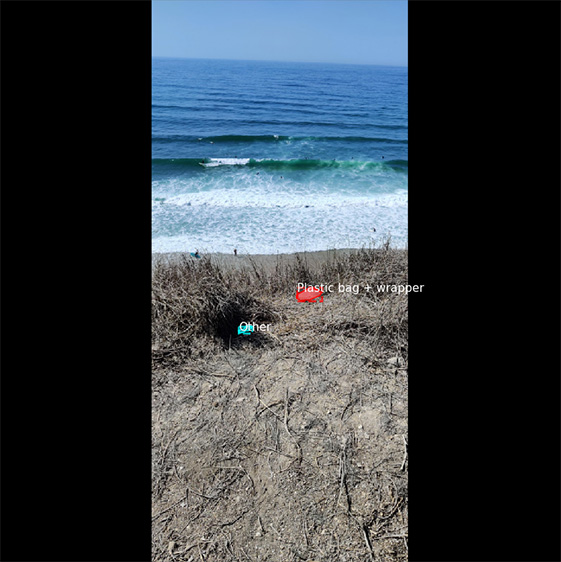}\\
		\includegraphics[width=0.245\linewidth]{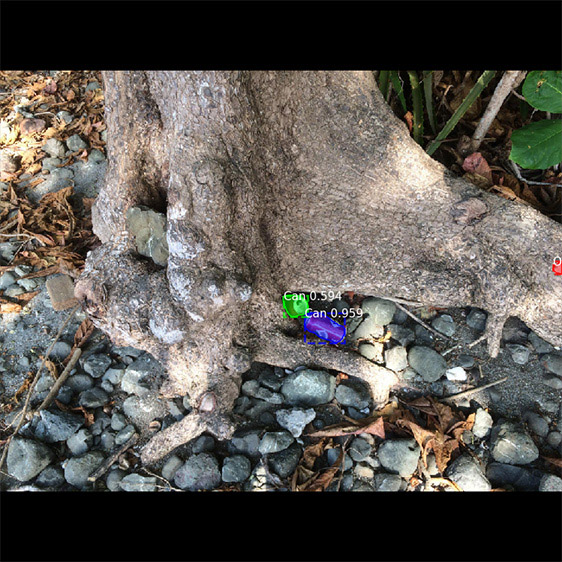}&
		\includegraphics[width=0.245\linewidth]{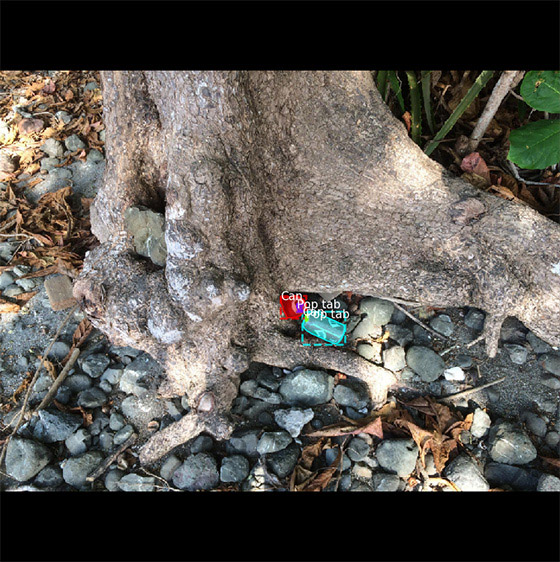}&
		\includegraphics[width=0.245\linewidth]{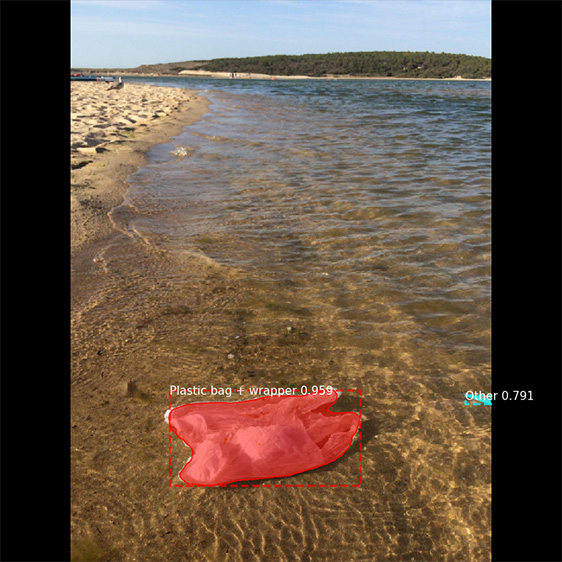}&
		\includegraphics[width=0.245\linewidth]{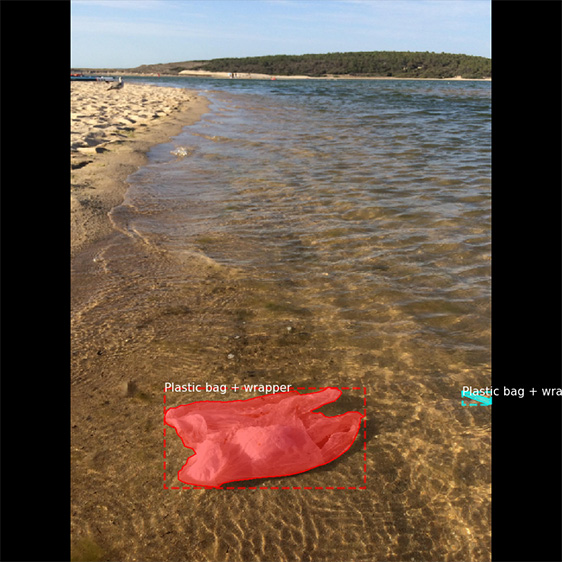}\\
		\includegraphics[width=0.245\linewidth]{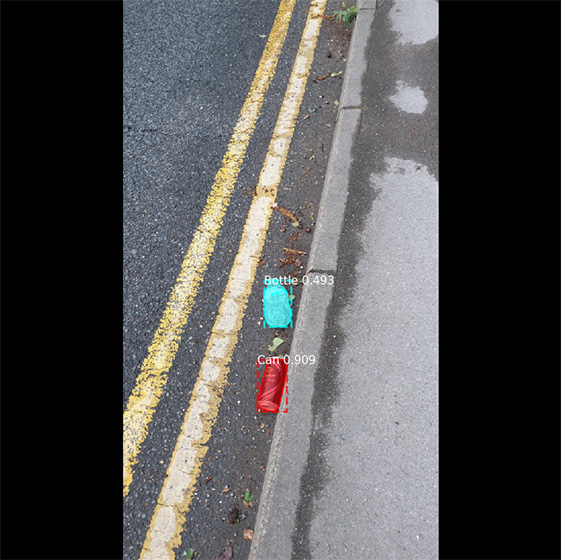}&
		\includegraphics[width=0.245\linewidth]{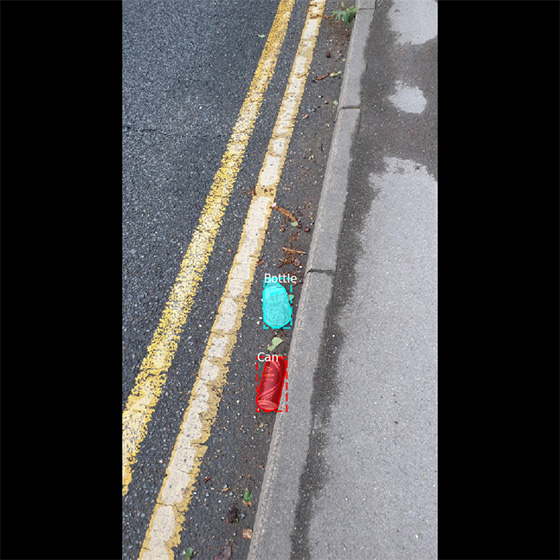}&
		\includegraphics[width=0.245\linewidth]{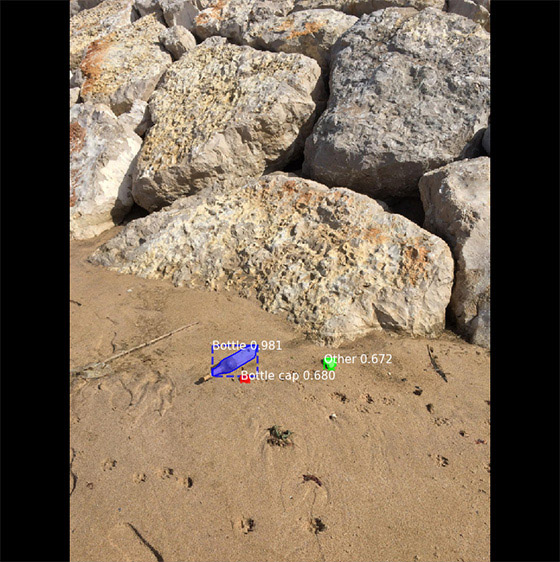}&
		\includegraphics[width=0.245\linewidth]{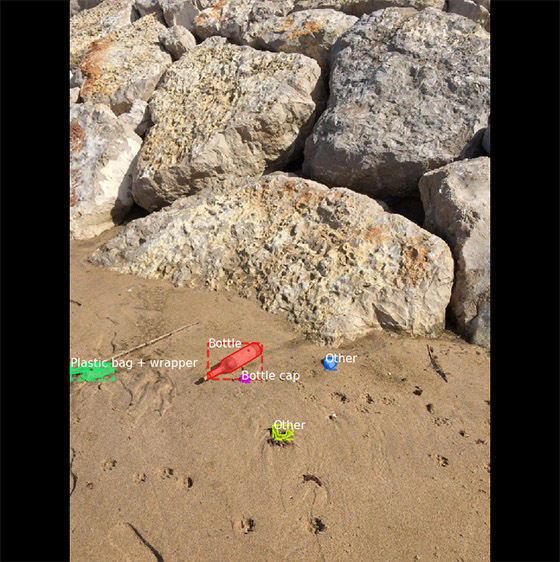}\\
	\end{tabular} 
	\caption{More examples of Mask R-CNN predictions on TACO-10 test-sets.}
	\label{fig:examples}
\end{figure*}

Several examples of detections on the test-sets are shown in Fig. \ref{fig:more_ex} and Fig. \ref{fig:examples}. The first image of Fig. \ref{fig:examples} shows that we are able to deal with transparent objects. However we see on the image below that seashells are picked as litter, namely \textit{other} and \textit{lid}. Overall, the performance is promising but the trained models are still prone to errors when faced with new or rare background, e.g., ocean waves, as shown by the third column and third row of Fig. \ref{fig:examples}.

\section{Conclusions and Future Work}

We showed how TACO dataset and tools can be used towards litter detection. We have received good feedback from many researchers, mostly undergrads and entrepreneurs who started working on this problem but struggled so far to find decent datasets that represent the problem. Although TACO is a good starting point, clearly, our dataset needs significantly more annotated images, thus we invite the reader to contribute annotations to TACO using our tools. \par
Moreover, detection results on tiny objects (e.g. cigarettes) using our network configuration is poor and affects significantly the overall AP, future work should devise better models and methods to fully exploit the high resolution of TACO images.  One could simply augment the input resolution but this increases significantly the memory footprint. Alternatively, one could run Mask R-CNN on a sliding window fashion and then fuse predictions, however this would sacrifice context from the surrounding windows. Thus, a more efficient and lossless method is required.

\bibliographystyle{ieeetr} 
{\footnotesize
\bibliography{ref}
}

\end{document}